\documentclass[table,final]{midl}
\usepackage{xcolor}
\usepackage{graphicx}
\usepackage{hyperref}
\usepackage{multirow}

\hypersetup{
    colorlinks=true,
    linkcolor=blue,
    filecolor=cyan,      
    urlcolor=magenta,
    }
\usepackage{booktabs}
\usepackage{natbib}
\newcommand{\transformer}{DAE-Former: Dual Attention-guided Efficient Transformer for Medical Image Segmentation}
\newcommand{\transformerlong}{DAE-Former}
\newcommand{\transformershort}{DAE-Former }
\newcommand{\githublink}{\href{https://github.com/mindflow-institue/DAEFormer}{GitHub}}
\newcommand{\mathdash}[1]{{\operatorname{#1}}}
\jmlrproceedings{PMLR}{}
\title[DAE-Former]{\transformer}

\midlauthor{\Name{Reza Azad\nametag{$^{1}$}} \Email{Reza.Azad@lfb.rwth-aachen.de}
\AND
\Name{René Arimond\nametag{$^{1}$}} \Email{Rene.Arimond@lfb.rwth-aachen.de}
\AND
\Name{Ehsan {Khodapanah Aghdam}\nametag{$^{2}$}} \Email{ehsan.khpaghdam@gmail.com}
\AND
\Name{Amirhossein Kazerouni\nametag{$^{3}$}} \Email{amirhossein477@gmail.com}
\AND
\Name{Dorit Merhof\nametag{$^{4,5}$}} \Email{dorit.merhof@ur.de}\\
\addr $^{1}$ Institute of Imaging and Computer Vision, RWTH Aachen University, Aachen, Germany\\
\addr $^{2}$ Department of Electrical Engineering, Shahid Beheshti University, Tehran, Iran\\
\addr $^{3}$ School of Electrical Engineering, Iran University of Science and Technology, Tehran, Iran\\
\addr $^{4}$ Institute of Image Analysis and Computer Vision, Faculty of Informatics and Data Science, University of Regensburg, Regensburg, Germany\\
\addr $^{5}$ Fraunhofer Institute for Digital Medicine MEVIS, Bremen, Germany
}

\begin{document}

\maketitle

\begin{abstract}
Transformers have recently gained attention in the computer vision domain due to their
ability to model long-range dependencies. However, the self-attention mechanism, which is the core part of the Transformer model, usually suffers from quadratic computational complexity with respect to the number of tokens. Many architectures attempt to reduce model complexity by limiting the self-attention mechanism to local regions or by redesigning the tokenization process. In this paper, we propose DAE-Former, a novel method that seeks to provide an alternative perspective by efficiently designing the self-attention mechanism. More specifically, we reformulate the self-attention mechanism to capture both spatial and channel relations across the whole feature dimension while staying computationally efficient. Furthermore, we redesign the skip connection path by including the cross-attention module to ensure the feature reusability and enhance the localization power. Our method outperforms state-of-the-art methods on multi-organ cardiac and skin lesion segmentation datasets without requiring pre-training weights. The code is publicly available at \githublink.

\end{abstract}


\section{Introduction}
Medical image segmentation has become one of the major challenges in computer vision. For physicians to monitor diseases accurately, visualize injuries, and select the correct treatment, stable and accurate image segmentation algorithms are necessary \cite{antonelli2022medical,kazerouni2022diffusion}. Deep learning networks perform very well in medical imaging, surpassing non-deep state-of-the-art (SOTA) methods.
However, an immense volume of data must be trained to achieve a good generalization with a large number of network parameters \cite{azad2022medical,bozorgpour2021multi,gupta2022segpc}. Additionally, the great need for large annotated data is another limitation of deep models, specifically in the medical domain. Unfortunately, per-pixel labeling is necessary for medical image segmentation, making annotation tedious and expensive \cite{aghdam2022attention,azad2020attention}.

A fully convolutional neural network (FCN) is one of the pioneering works utilized for image segmentation. To preserve the spatial information and reconstruct the segmentation map, the FCN network uses convolutional layers on both encoder and decoder modules without any fully-connection, resulting in a less parametrized model and a better generalization performance. The U-Net by Ronneberger et al. \cite{ronneberger2015unet} further enhances the FCN architecture by designing skip connections in each scale of the encoder/decoder modules. Unlike FCN, U-Net performs segmentation tasks well without requiring huge annotated training sets. Skip connections can therefore be seen as the key component for the U-Net's success \cite{azad2022unetreview}. 
Several extensions to the U-Net have been proposed to enhance the performance of the network \cite{zhou2018unet++,huang2020unet,valanarasu2020kiu}.
These methods aim to enrich the feature representation either by incorporating the attention mechanisms \cite{oktay2018attention,SEnet,woo2018cbam}, or redesigning the skip connection path \cite{zhou2018unet++,azad2019bi}, or replacing the backbone module~\cite{karaali2022dr}. Although these extensions improve the feature representation, the locality restriction of the convolution layer limits the representational power of these networks to capture the shape and structural information, which is crucial for medical image segmentation. It has been shown that exploiting shape information in CNNs by fine-tuning the input images can boost the representational power of the network \cite{azad2021deep}. However, including shape representations inside the CNN networks requires modeling long-range dependencies in the latent space, which is still an open challenge in CNN architectures \cite{azad2021texture,feyjie2020semi}. 

To address CNN limitation, the Vision Transformer (ViT) \cite{dosovitskiy2020vit} model has been proposed. The ViT architecture is purely based on the multi-head self-attention mechanism. This enables the network to capture long-range dependencies and encode shape representations. ViT, however, requires large amounts of training data to perform similarly to CNNs, and the self-attention mechanism suffers from a quadratic computation complexity with respect to the number of tokens. The naive ViT model also renders poor performance compared to the CNN model for capturing the local representation. To address the weak local representation of the Transformer model, several methods have been proposed to build a hybrid CNN-Transformer network \cite{chen2021transunet,heidari2022hiformer}. TransUNet \cite{chen2021transunet}, as a pioneer work in this direction, offers a hierarchical Transformer that captures global and fine-grained local context by combining convolutions and an attention mechanism. However, the downside of TransUNet is its high number of parameters and computational inefficiency. Additionally, although Hiformer \cite{heidari2022hiformer} and contextual network \cite{reza2022contextual} effectively bridge a CNN and a Transformer for medical image segmentation, it still relies on a heavy CNN backbone.

To address the computational complexity of the Transformer model, recent designs suggest either imposing a restriction on the self-attention mechanism to perform in a local region \cite{ding2022davit, liu2021swin, chen2021regionvit}, or defining a scaling factor to reduce the spatial dimension \cite{huang2021missformer}, or calculating channel attention instead of spatial attention \cite{ali2021xcit}. However, the global context is only partially captured in such methods.
Swin-Unet \cite{cao2021swin} takes a different perspective and offers a U-Net-like pure Transformer architecture that operates at different scales and fuses the features from different layers using skip connections. Swin-Unet uses two consecutive Transformer blocks with different windowing settings (shifted windows to reduce the computational burden) to attempt to
recapture context from neighboring windows. Although the multi-scale representation of the Swin-Unet enhances the feature representation, the spatial context is still limited in the process.

To address the aforementioned limitations, we propose a dual attention module that operates on the full spatial dimension of the input feature, and also captures the channel context. To this end,  we apply efficient attention by Shen et al. \cite{shen2021efficient}, which reduces the complexity of self-attention to linear while producing the same output as the regular self-attention mechanism. Moreover, 
to capture the input feature's channel context, we reformulate the attention mechanism with the cross-covariance method \cite{ali2021xcit}. 
We integrate our redesigned Transformer block in a hierarchical U-Net-like pure Transformer architecture, namely, the \transformerlong. In order to reliably fuse multi-scale features from different layers, we propose a cross-attention module in each skip connection path. Our contributions are as follows: \textbf{1)} a novel efficient dual attention mechanism to capture the full spatial and channel context of the input feature vector, \textbf{2)} a skip connection cross attention (SCCA) module to adaptively fuse features from encoder and decoder layers, and \textbf{3)} a hierarchical U-Net-like pure Transformer structure for medical image segmentation.


\section{Proposed Method}\label{sec:method}
\begin{figure}[!t]
    \includegraphics[width=\textwidth]{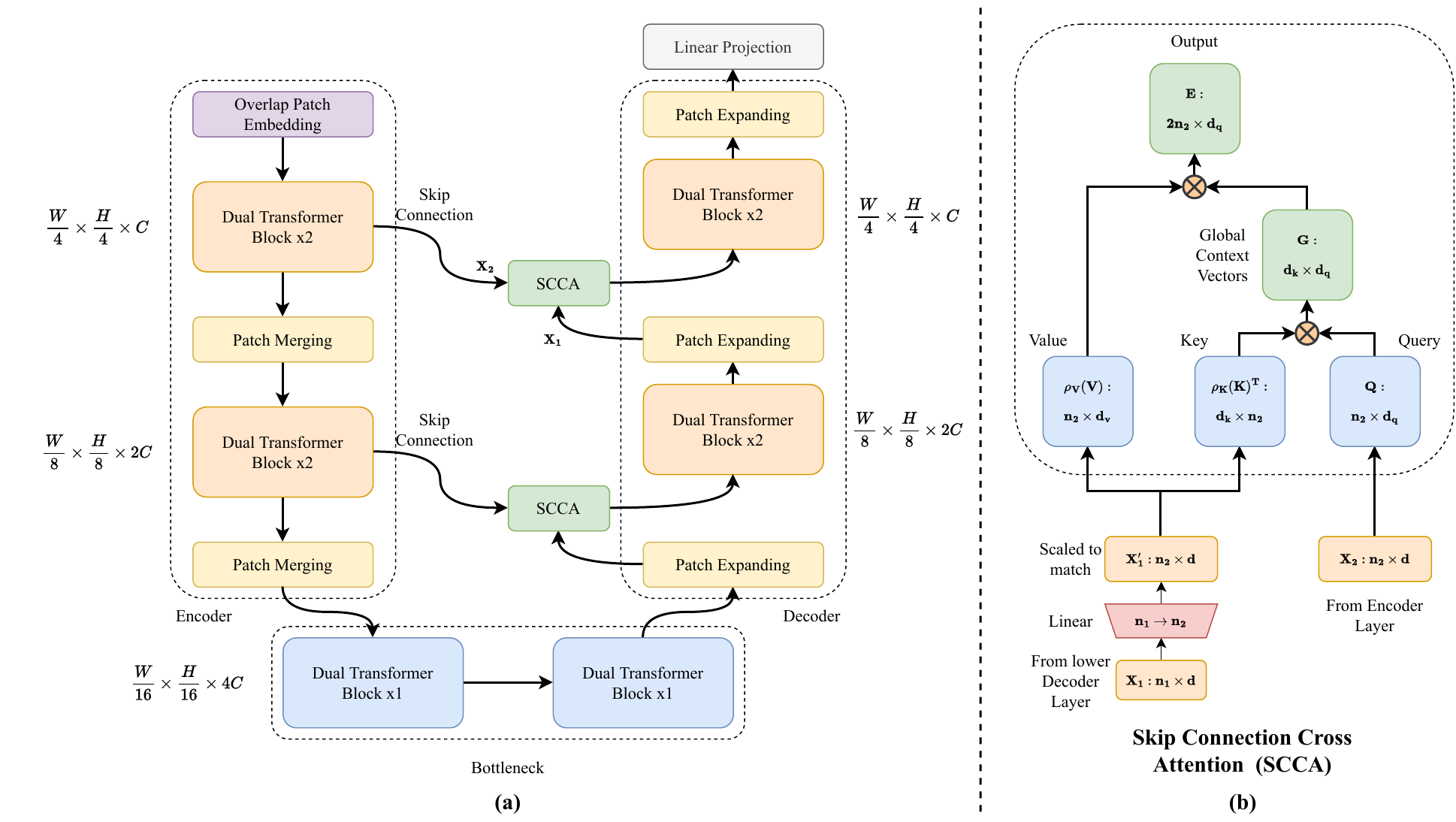}
    \caption{(a): The structure of our~\transformerlong. Both the encoder and decoder of the U-Net-like architecture are each comprised of 3 blocks. Our dual attention block consists of efficient attention followed by transpose attention. (b): The skip connection cross attention (SCCA) fuses information from the encoder layers with the features from the lower decoder layer.}
    \label{fig:proposed_method}
\end{figure}

We introduce the \transformerlong~(\figureref{fig:proposed_method}), a convolution-free U-Net-like hierarchical pure Transformer. Given an input image \textbf{$x^{H\times W \times C}$} with spatial dimension $H \times W$ and $C$ channels, the \transformerlong~utilizes the patch embedding module~\cite{cao2021swin, huang2021missformer} to gain overlapping patch tokens of size $4 \times 4$ from the input image. The tokenized input ($x^{n \times d}$) then goes through the encoder module, with 3 stacked encoder blocks, each consisting of two consecutive dual Transformer layers and a patch merging layer. During patch merging, $2 \times 2$ patch tokens are merged to reduce the spatial dimension while doubling the channel dimension.
This allows the network to gain a multi-scale representation in a hierarchical fashion. In the decoder, the tokens are expanded again by a factor of 2 in each block. The output of each patch expanding layer is then fused with the features forwarded by the skip connection from the parallel encoder layer using SCCA.
The resulting features are fed into two consecutive dual Transformer layers. Finally, a linear projection layer produces the output segmentation map. In the next sections, we will first provide a brief overview of the efficient and transpose attentions. Then we will introduce our efficient dual attention and the SCCA modules.

\subsection{Efficient Attention}
The standard self-attention mechanism~\eqref{eq:self_attention} suffers from a quadratic computational complexity ($O(N^2)$), which limits the applicability of this architecture for high-resolution images. Q, K, and V in ~\eqref{eq:self_attention} shows the query, key, and value vectors and $d$ is the embedding dimension,

\begin{equation}
S(\mathbf{Q}, \mathbf{K}, \mathbf{V}) =\operatorname{softmax}\left(\frac{\mathbf{Q K^T}}{\sqrt{d_{\mathbf{k}}}}\right) \mathbf{V}.\label{eq:self_attention}
\end{equation}

Efficient attention by Shen et al.~\cite{shen2021efficient} uses the fact that regular self-attention produces a redundant context matrix to propose an efficient way to compute the self-attention procedure \eqref{eq:efficient-attention}:
\begin{align}
    \label{eq:efficient-attention}
    \mathbf{E}(\mathbf{Q},\mathbf{K},\mathbf{V}) =  \mathbf{\rho_{q}}(\mathbf{Q})(\mathbf{\rho_{k}}(\mathbf{K})^{\mathbf{T}}\mathbf{V}),
\end{align}
where $\rho_{q}$ and $\rho_{k}$ are normalization functions for the queries and keys. It has been shown in \cite{shen2021efficient} that the module produces an equivalent output of dot-product attention when $\rho_{q}$ and $\rho_{k}$ are applied, which are softmax normalization functions. Therefore, efficient attention normalizes the keys and queries first, then multiplies the keys and values, and finally, the resulting global context vectors are multiplied by the queries to produce the new representation.

Unlike dot-product attention, efficient attention does not  first compute pairwise similarities between points. Instead, the keys are represented as $d_{k}$ attention maps $\mathbf{k^{T}}_{j}$, with $j$ referring to position $j$ in the input feature.
These global attention maps represent a semantic aspect of the whole input feature instead of similarities to the position of the input.
This shifting of orders drastically reduces the computational complexity of the attention mechanism while maintaining high representational power.
The memory complexity of efficient attention is $O(dn + d^{2})$
while the computational complexity is $O(d^{2}n)$ when $d_{v} = d, d_{k} = \frac{d}{2}$ - which is a typical setting. In our structure, we use efficient attention to capture the spatial importance of the input feature map.

\subsection{Transpose Attention}
Cross-covariance attention, also called transpose attention~\cite{ali2021xcit}, is a channel attention mechanism. This strategy employs transpose attention only to enable the processing of larger input sizes. However, we reformulate the problem and propose a transpose attention mechanism to capture the full channel dimension efficiently. The transpose attention is shown in Equation \eqref{eq:transpose-attention}:

\begin{align}
    \label{eq:transpose-attention}
    \mathbf{T}(\mathbf{Q},\mathbf{K},\mathbf{V}) = \mathbf{V} \mathcal{C}_{T}(\mathbf{K},\mathbf{Q}), \quad\mathcal{C}_{T}(\mathbf{K},\mathbf{Q}) = Softmax(\mathbf{K}^{\mathbf{T}}\mathbf{Q}/\tau)
\end{align}

The keys and queries are transposed, and, therefore, the attention weights are based on the cross-covariance matrix. $\mathcal{C}_{T}$ refers to the context vector of the transpose attention. The temperature parameter $\tau$ is introduced to counteract the scaling with the $l_{2}$-norm that is applied to the queries and keys before calculating the attention weights. This increases stability during training but removes a degree of freedom, thus reducing the representational power of the module.

Transpose attention has a time complexity of $O(Nd^{2}/h)$, whereas standard self-attention requires $O(N^{2}d)$. The space
complexity is $O(hN^{2}+Nd)$ for transpose attention and $O(d^{2}/h+Nd)$ for self-attention. Self-attention scales quadratically with the number of tokens $N$, whereas transpose attention scales quadratically with the embedding dimension $d$, which is usually smaller than $N$, especially for larger images.

\subsection{Efficient Dual Attention}
\begin{figure}[h]
    \centering
    \includegraphics[width=0.7\textwidth]{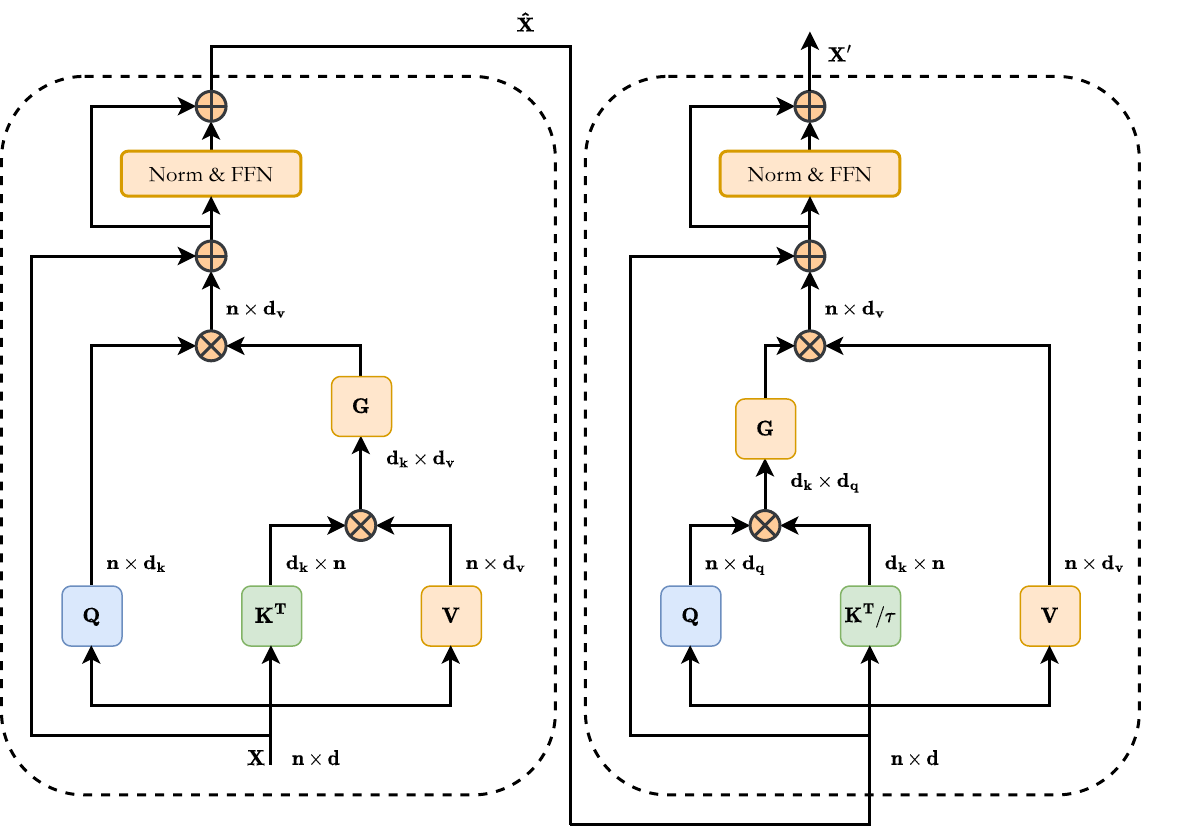}
    \caption{The efficient dual attention block. It consists of an efficient attention block, followed by a Norm \& FFN, and a channel attention block followed by a Norm \& FFN to perform spatial and channel attentions.}
    \label{fig:dual_attention}
    \vspace{-2em}
\end{figure}

A literature review \cite{guo2022attention} on the attention mechanism shows that combining spatial and channel attention enhances the capacity of the model to capture more contextual features than single attention. Therefore, we construct a dual Transformer block that combines transpose (channel) attention and efficient (spatial) attention. The structure of our efficient dual attention block is shown in~\figureref{fig:dual_attention}.
Our efficient dual attention block \eqref{eq:dual_attention} consists of an efficient attention \eqref{eq:e_block}, followed by an add \& norm~\eqref{eq:mlp_1}, and a transpose attention block that performs the channel attention \eqref{eq:t_block}, followed by an add \& norm \eqref{eq:mlp_2}.

\begin{align}
    \mathbf{E_{block}}(\mathbf{X}, \mathbf{Q_{1}}, \mathbf{K_{1}}, \mathbf{V_{1}})         & = \mathbf{E}(\mathbf{Q_{1}},\mathbf{K_{1}},\mathbf{V_{1}}) + \mathbf{X} \label{eq:e_block},                                                    \\
    \operatorname{MLP}_{1}(\mathbf{E_{block}})                                             & = \operatorname{MLP}(\operatorname{LN}(\mathbf{E_{block}})) \label{eq:mlp_1},                                                                \\
    \mathbf{T_{block}}(\mathbf{E_{block}}, \mathbf{Q_{2}}, \mathbf{K_{2}}, \mathbf{V_{2}}) & = \mathbf{T}(\operatorname{MLP}_{1}(\mathbf{E_{block}}) + \mathbf{E_{block}}) + \operatorname{MLP}_{1}(\mathbf{E_{block}}) \label{eq:t_block}, \\
    \operatorname{MLP}_{2}(\mathbf{T_{block}})                                             & = \operatorname{MLP}(\operatorname{LN}(\mathbf{T_{block}})) \label{eq:mlp_2},                                                                \\
    \operatorname{DualAttention}(\mathbf{T_{block}})                                       & = \operatorname{MLP}_{2}(\mathbf{T_{block}}) + \mathbf{T_{block}} \label{eq:dual_attention}.
\end{align}
$\mathbf{E(\cdot)}, \mathbf{T(\cdot)}$ refer to efficient attention and transpose attention, respectively. $\mathbf{T_{block}}$ denotes the transpose attention block and $\mathbf{E_{block}}$ the efficient attention block.
$\mathbf{Q_{1}},\mathbf{K_{1}},\mathbf{V_{1}}$ are the keys, queries, and values calculated from the input feature $\mathbf{X}$.~$\mathbf{Q_{2}},\mathbf{K_{2}},\mathbf{V_{2}}$ are
the queries, keys, and values computed from the input to the transpose attention block. $MLP$ denotes the $\mathdash{Mix-FFN}$ feed-forward network~\cite{huang2021missformer}:

\begin{align}
    \operatorname{MLP}(\mathbf{X}) = \operatorname{FC}(\operatorname{GELU}(\mathdash{DW-Conv}(\operatorname{FC}(\mathbf{X}))))
\end{align}
with $\operatorname{FC}$ being a fully connected layer. $\operatorname{GELU}$ refers to GELU activation~\cite{hendrycks2016gelu} and $\mathdash{DW-Conv}$ is depth-wise convolution.

\subsection{Skip Connection Cross Attention}

The SCCA module is shown in~\figureref{fig:proposed_method}(b).
Instead of simply concatenating the features from the encoder and decoder layers, we cross-attend them to preserve the underlying features more efficiently.
Our proposed module can effectively provide spatial information to each decoder so that it can recover fine-grained details when producing output masks. In our structure, the skip connection cross attention (SCCA) applies efficient attention, but, instead of using the same input feature for keys, queries, and values, the input used for the query is the output of the encoder layer forwarded by a skip connection $X_{2}$, hence the name.
The input used for keys and values is the output of the lower decoder layer $X_{1}$.
To fuse the two features, $X_{1}$ needs to be scaled to the same embedding dimension as $X_{2}$ using a linear layer \eqref{eq:scca-attention}. The motivation behind using $X_{2}$ as an input for the query is to model the multi-level representation within the efficient attention block.

\begin{align}
    \label{eq:scca-attention}
    \mathbf{X'_{1}} & = FC(X_{1}), \quad\mathbf{K, V}  = Proj(\mathbf{X'_{1}}), \quad\mathbf{Q}       = Proj(\mathbf{X_{2}}) \nonumber \\
    \mathbf{E}      & = \mathbf{\rho_{v}}(\mathbf{V})\mathbf{\rho_{k}}(\mathbf{K^{T}})\mathbf{Q}.
\end{align}
Here, $\rho_{v}, \rho_{k}$ are normalization functions, and $Proj$ refers to a projection function. In this case, it is a linear projection.

\section{Experimental Setup}
Our proposed method is implemented in an end-to-end manner using the PyTorch library and is trained on a single RTX 3090 GPU. The training is done with a batch size of 24 and a stochastic gradient descent with a base learning rate of 0.05, a momentum of 0.9, and a weight decay of 0.0001. The model is trained for 400 epochs using both cross-entropy and Dice losses ($Loss = 0.6 \cdot L_{dice} + 0.4 \cdot L_{ce}$).

\subsection{Dataset and Evaluation Metrics}
We use the publicly available Synapse dataset for the evaluation process, which constitutes a multi-organ segmentation dataset with 30 cases with 3779 axial abdominal clinical CT images~\cite{cao2021swin, huang2021missformer}. We follow the setting presented in \cite{chen2021transunet} for the evaluation. We further evaluate our method on the skin lesion segmentation challenge using the ISIC 2018 \cite{codella2019skin} dataset. We follow \cite{azad2022contextual} for the experimental setup.

\subsection{Quantitative and Qualitative Results}
\begin{figure}
    \centering
    \includegraphics[width=1\textwidth]{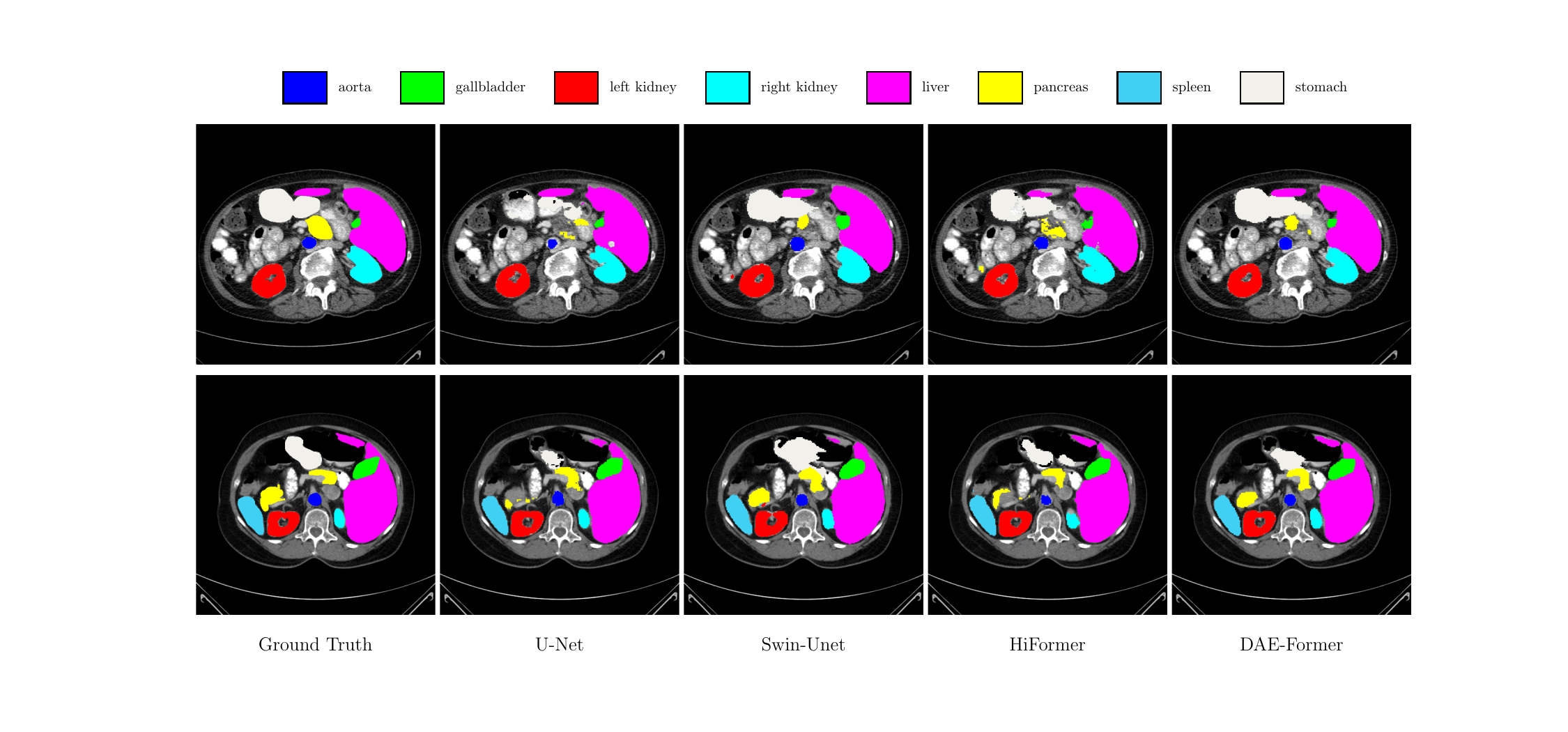}
    \caption{Comparative segmentation results on the \textit{Synapse} dataset.}
    \label{fig:synapsevisualization}
\end{figure}

\tableref{tab:performance_comparison} presents the performance of our proposed \transformershort on the Synapse dataset.
\transformershort surpasses the previous state-of-the-art (SOTA) methods in terms of DSC score. It also outperforms CNN-based methods by a large margin.
We confirm an increase of the Dice score by 0.67\% compared to the previous state-of-the-art, MISSFormer~\cite{huang2021missformer}. 
We observe that the segmentation performance increases, especially for the gallbladder, kidney, liver, and spleen. \figureref{fig:synapsevisualization} shows the visualization of the segmentation maps. It can be observed that the competitive methods fail to predict the small organs (e.g., pancreas) while our model produces a smooth segmentation map for all organs.

\begin{table*}[!b]
    \centering
    \caption{Comparison results of the proposed method on the \textit{Synapse} dataset. \textcolor{blue}{Blue} indicates the best result, and  \textcolor{red}{red} indicates the second-best. We exclusively report each method's parameter numbers in order of millions (M).}
    
    \resizebox{1\textwidth}{!}{
        \begin{tabular}{l|c|cc|cccccccc}
            \toprule
            \textbf{Methods}               & \textbf{\# Params (M)}       & \textbf{DSC~$\uparrow$} & \textbf{HD~$\downarrow$} & \textbf{Aorta}          & \textbf{Gallbladder}    & \textbf{Kidney(L)}      & \textbf{Kidney(R)}      & \textbf{Liver}          & \textbf{Pancreas}       & \textbf{Spleen}         & \textbf{Stomach}        \\
            \midrule
            U-Net \cite{ronneberger2015unet}     &  14.8    & 76.85                   & 39.70                    & \textcolor{red}{89.07}  & \textcolor{red}{69.72}  & 77.77                   & 68.60                   & 93.43                   & 53.98                   & 86.67                   & 75.58
            \\
            Att-UNet \cite{schlemper2019attention} &  34.88  & 77.77                   & 36.02                    & \textcolor{blue}{89.55} & 68.88                   & 77.98                   & 71.11                   & 93.57                   & 58.04                   & 87.30                   & 75.75
            \\
            TransUNet \cite{chen2021transunet}     &  105.28 & 77.48                   & 31.69                    & 87.23                   & 63.13                   & 81.87                   & 77.02                   & 94.08                   & 55.86                   & 85.08                   & 75.62
            \\
            Swin-Unet \cite{cao2021swin}       & 27.17     & 79.13                   & 21.55                    & 85.47                   & 66.53                   & 83.28                   & 79.61                   & 94.29                   & 56.58                   & 90.66                   & 76.60
            \\
            LeVit-Unet \cite{xu2021levit}  & 52.17       & 78.53                   & 16.84                    & 78.53                   & 62.23                   & 84.61                   & 80.25                   & 93.11                   & 59.07                   & 88.86                   & 72.76
            \\
            MT-UNet \cite{wang2022mixed} & 79.07 & 78.59 & 26.59 & 87.92 & 64.99 & 81.47 & 77.29 & 93.06 & 59.46 & 87.75 & 76.81 \\
            TransDeepLab \cite{azad2022transdeeplab} & 21.14 & 80.16 & 21.25 & 86.04 & 69.16 &84.08 & 79.88 & 93.53 & 61.19 & 89.00 & 78.40 \\
            HiFormer \cite{heidari2022hiformer}   & 25.51 & 80.39  & \textcolor{blue}{14.70}  & 86.21  & 65.69  & \textcolor{red}{85.23}  & 79.77 & 94.61 & 59.52  & 90.99 & \textcolor{blue}{81.08}
            \\
            MISSFormer \cite{huang2021missformer} & 42.46 & \textcolor{red}{81.96}  & 18.20 & 86.99 & 68.65 & 85.21 & \textcolor{red}{82.00} & 94.41 & \textcolor{blue}{65.67} & \textcolor{blue}{91.92}  & \textcolor{red}{80.81}
            \\
            \hline
            EffFormer (baseline model)           &    22.31              & 80.79                   & 17.00   & 85.81                   & 66.89                   & 84.10                   & 81.81 & \textcolor{red}{94.80}  & 62.25                   & 91.05                   & 79.58
            \\
            DAE-Former (without SCCA) &  40.75 & 81.59 & 17.31  & 87.41  & 69.57  & 85.22 & 80.46 & 94.68 & 63.71  & 91.47 & 78.23
            \\
            \rowcolor{cyan!10}
            \textbf{\transformershort}      &   48.01   & \textcolor{blue}{82.63} & \textcolor{red}{16.39} & 87.84 & \textcolor{blue}{71.65} & \textcolor{blue}{87.66} & \textcolor{blue}{82.39}  & \textcolor{blue}{95.08} & \textcolor{red}{63.93}  & \textcolor{red}{91.82} & 80.77
            \\
            \bottomrule
        \end{tabular}
    }\label{tab:performance_comparison}
\end{table*}

The results on the skin lesion segmentation benchmarks are presented in~\tableref{tab:quantitative}. It can be seen that, compared to both CNN and Transformer-based methods, our network produces better performance in almost all metrics. Compared to the most competitive and similar approach (TMU-Net), our model performs better. The TMU-Net requires extra information (boundary and foreground distribution information) and suffers from a large number of parameters (approximately 165.1M vs. 48.1M parameters for our model). We also observe that the dual attention mechanism performs better than the base structure, indicating our suggested modules' effectiveness for better performance gain. In addition, qualitative results are depicted in Appendix \ref{skin-results}.

\begin{table}[!tbh]
    \centering
    \caption{Performance comparison of the proposed method against the SOTA approaches on the \textit{ISIC2018} skin lesion segmentation task.} \label{tab:quantitative}
    \resizebox{0.8\textwidth}{0.1\textheight}{
        \begin{tabular}{l || c c c c }
            \hline
            \textbf{Methods}                    & \textbf{DSC}    & \textbf{SE}     & \textbf{SP}     & \textbf{ACC}    \\
            \hline
            U-Net \cite{ronneberger2015unet}       & 0.8545          & 0.8800          & 0.9697          & 0.9404          \\
            Att U-Net \cite{oktay2018attention} & 0.8566          & 0.8674          & \textbf{0.9863} & 0.9376          \\
            TransUNet \cite{chen2021transunet}  & 0.8499          & 0.8578          & 0.9653          & 0.9452          \\
            MCGU-Net \cite{asadi2020multi}      & 0.895           & 0.848           & 0.986           & 0.955           \\
            MedT \cite{valanarasu2021medical}   & 0.8389          & 0.8252          & 0.9637          & 0.9358          \\
            FAT-Net \cite{wu2022fat}            & 0.8903          & 0.9100 & 0.9699          & 0.9578          \\
            TMU-Net \cite{azad2022contextual}   & 0.9059          & 0.9038          & 0.9746          & 0.9603          \\
            Swin\,U-Net \cite{cao2021swin}      & 0.8946          & 0.9056          & 0.9798          & 0.9645          \\
            TransNorm \cite{azad2022transnorm}  & 0.8951          & 0.8750          & 0.9790          & 0.9580          \\
            \hline
            EffFormer  & 0.8904 & 0.8861 & 0.9698 & 0.9519  \\
            \transformershort (without SCCA) & 0.8962& 0.8634&	0.9830& 0.9578          \\
            \rowcolor{cyan!10}
            \textbf{\transformershort} & \textbf{0.9147} & \textbf{0.9120} & 0.9780 & \textbf{0.9641}          \\ \hline
        \end{tabular}
    }
\end{table}

\begin{figure}[!thb]
    \centering
    \includegraphics[width=\textwidth]{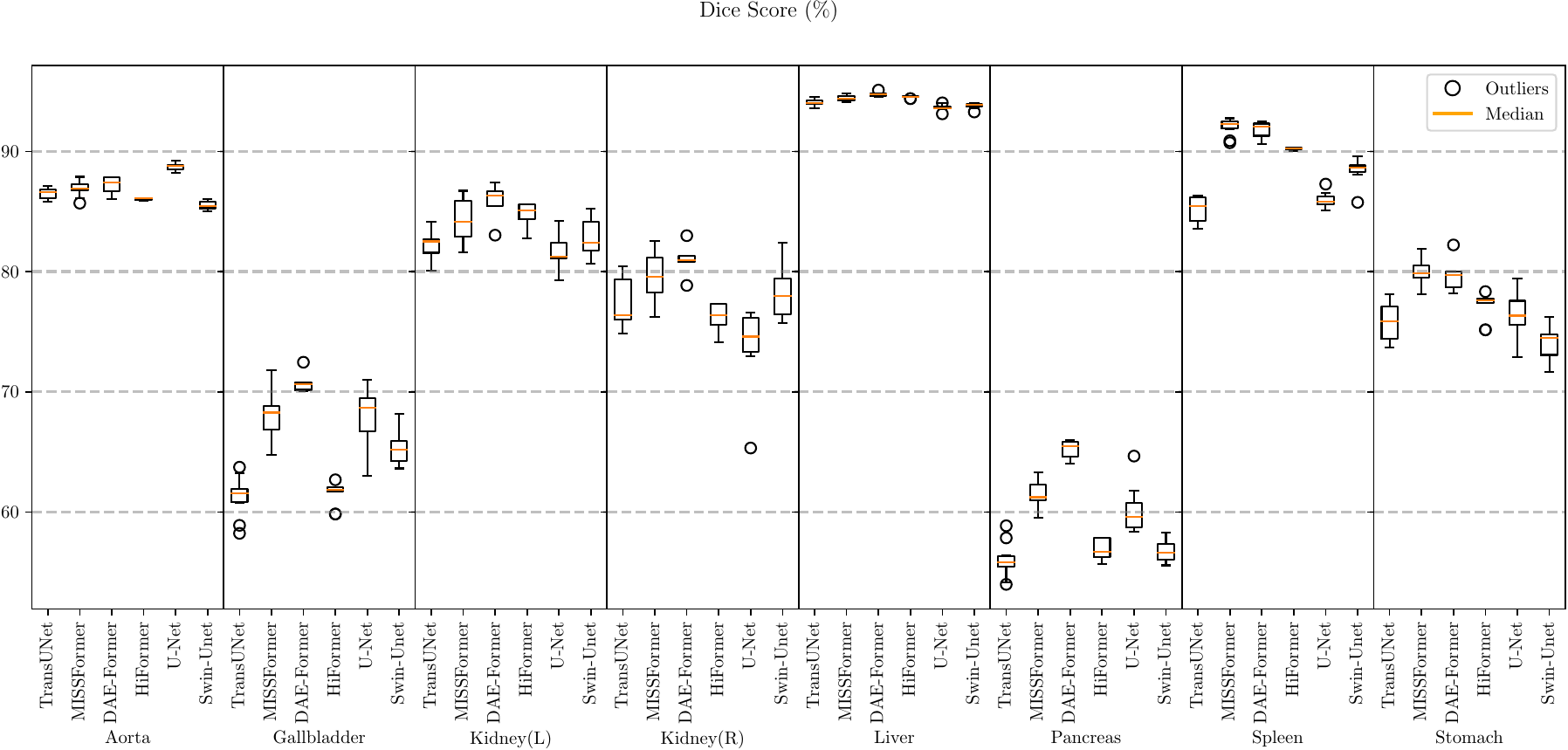}
    \caption{Statistical analysis between the U-Net \cite{ronneberger2015unet}, TransUNet \cite{chen2021transunet}, Swin-Unet \cite{cao2021swin}, HiFormer \cite{heidari2022hiformer}, MISSFormer \cite{huang2021missformer}, and our proposed method, DAE-Former.}
    \label{fig:statistical_analysis}
\end{figure}

\begin{figure}[!thb]
\centering
\resizebox{1\textwidth}{!}{
    \begin{tabular}{@{} *{6}c @{}}
    \includegraphics[width=0.25\textwidth]{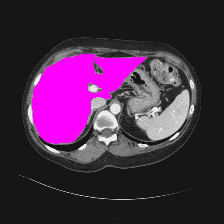} &
    \includegraphics[width=0.25\textwidth]{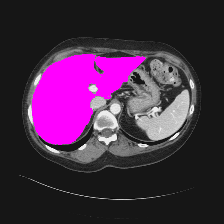} &
    \includegraphics[width=0.25\textwidth]{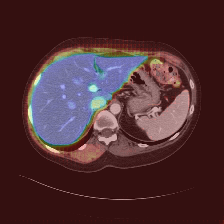} & 
    \includegraphics[width=0.25\textwidth]{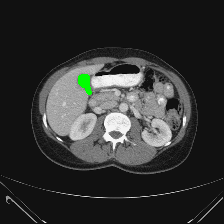} &
    \includegraphics[width=0.25\textwidth]{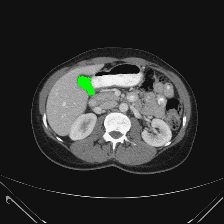} &
    \includegraphics[width=0.25\textwidth]{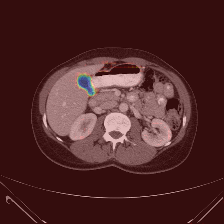} \\
    \includegraphics[width=0.25\textwidth]{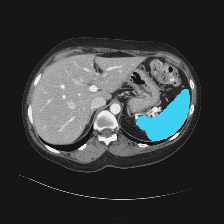} &
    \includegraphics[width=0.25\textwidth]{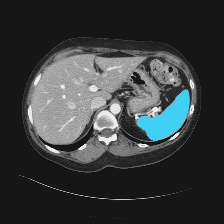} &
    \includegraphics[width=0.25\textwidth]{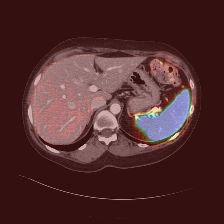} &
    \includegraphics[width=0.25\textwidth]{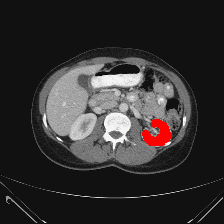} &
    \includegraphics[width=0.25\textwidth]{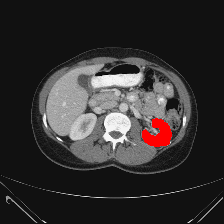} &
    \includegraphics[width=0.25\textwidth]{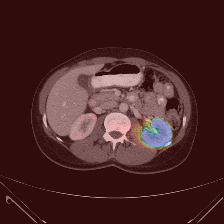} \\
    \includegraphics[width=0.25\textwidth]{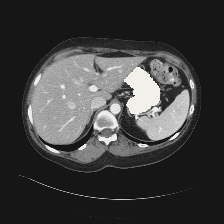} &
    \includegraphics[width=0.25\textwidth]{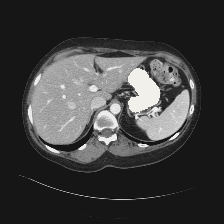} &
    \includegraphics[width=0.25\textwidth]{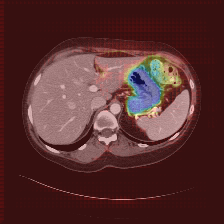} &
    \includegraphics[width=0.25\textwidth]{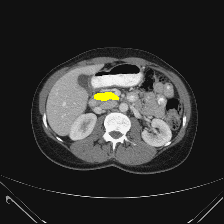} &
    \includegraphics[width=0.25\textwidth]{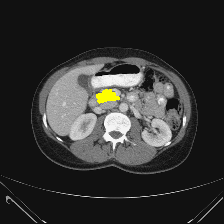} &
    \includegraphics[width=0.25\textwidth]{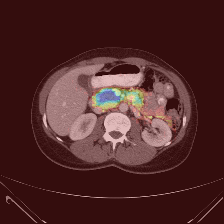} \\
    {\small (a) Ground Truth} & {\small(b) Prediction} & {\small(c) Heatmap} & {\small (d) Ground Truth} & {\small(e) Prediction} & {\small(f) Heatmap}
    \end{tabular}
}
\caption{Visualization of attention maps using Grad-CAM on the \textit{Synapse} dataset demonstrates the effectiveness of DAE-Former in detecting both large (liver, spleen, and stomach) and small (gallbladder, left kidney, and pancreas) organs.} \label{fig:largeorgan}
\end{figure}

To endorse the validity of the results we conducted a statistical analysis by running 10 times SOTA and our proposed models to report the mean Dice score over each organ. \figureref{fig:statistical_analysis} indicates that the DAE-Former performs with low variance and confident performances over Gallbladder, Kidney (L/R), and liver. We also provide attention map visualization (\figureref{fig:largeorgan}) of the DAE-Former using Grad-CAM to highlight its capacity for capturing local and global dependency. 

\section{Conclusion}
In this paper, we propose \transformerlong, a novel U-Net-like hierarchical pure Transformer that leverages both spatial and channel attention on the full feature dimension. We enrich the representational space by including dual attention while retaining the same number of parameters compared to previous architectures. Furthermore, we perform a fusion of multi-scale features through skip connection cross-attention. Our model achieves the SOTA results on both the Synapse and skin lesion segmentation datasets, thereby surpassing CNN-based approaches by a large margin.

\bibliography{refs}
\let\cleardoublepage\clearpage

\newpage
\appendix

\section{Attention in Detail}
\begin{figure}[h]
    \includegraphics[width=\textwidth]{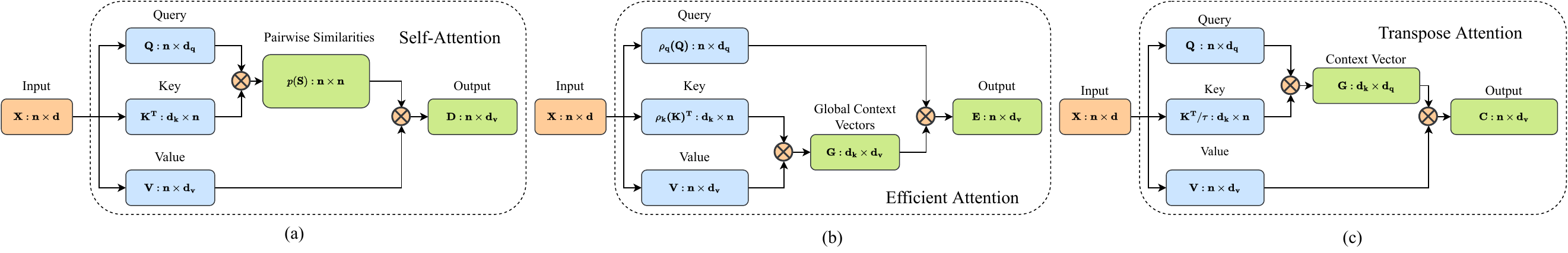}
    \caption{(a) Standard dot-product attention, (b) efficient attention~\cite{shen2021efficient}, and (c) transpose attention~\cite{ali2021xcit}. Redrawn for comparative overview.}\label{fig:eff_scca}
\end{figure}

In this section, we elaborate more on our proposed method's attention mechanism and design choice. Initially, we visualize the standard self-attention, efficient self-attention~\cite{shen2021efficient} and transposed self-attention~\cite{ali2021xcit} modules in \figureref{fig:eff_scca}. The standard self-attention performs the calculation based on the tokens and is, therefore, not an ideal choice for high-resolution images. The efficient attention module, however, first multiplies the keys and values to obtain a global context vector. The resulting calculation requires the keys and queries to be normalized beforehand to obtain the same result as self-attention, but the computational complexity is linear with respect to the number of tokens. Transpose attention (\figureref{fig:eff_scca}(c)) is also linear, but due to the keys being transposed, the attention is calculated on the channel dimension of the input tensor.

In our method, we take into account the computational advantages of the efficient and the transpose attention mechanisms to build a dual attention mechanism (channel and spatial attentions as presented in the main text). There are several ways in which the channel and spatial attention blocks may be combined.
The first option is to employ sequential dual attention. Here, the channel attention block is applied to the output of the efficient spatial attention block or vice versa.
The two attention blocks can be computed in parallel on the same input. The outputs of both blocks must be combined, which presents more options:
The tensors can either be added, which is referred to as ``additive dual attention,'' or concatenated and fed to an MLP, which reduces the dimension to the input dimension.

\begin{figure}[h]
  \begin{subfigure}[]
    \centering
    \includegraphics[width=0.45\linewidth]{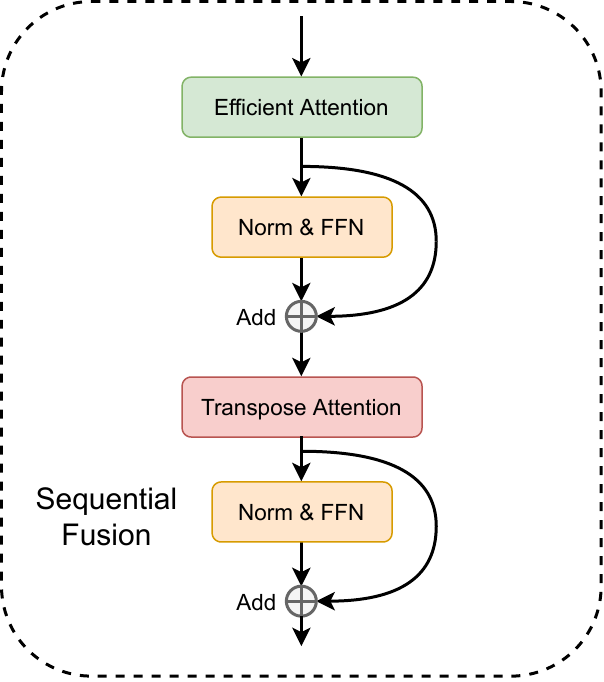}
  \end{subfigure}
  \begin{subfigure}
    \centering
    \includegraphics[width=0.45\linewidth]{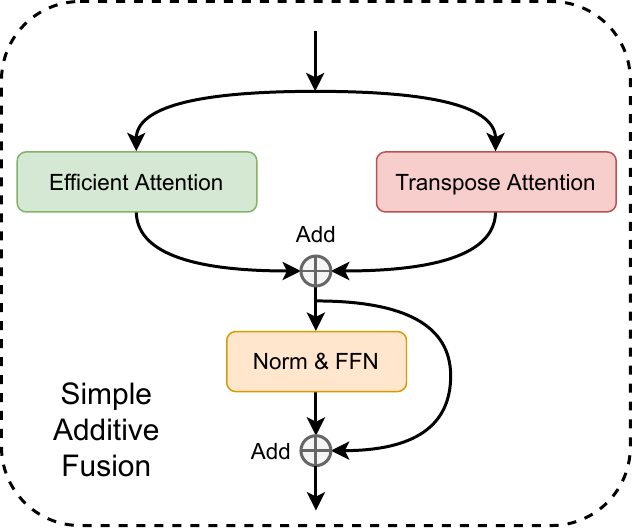}
  \end{subfigure}
  \begin{subfigure}
    \centering
    \includegraphics[width=0.45\linewidth]{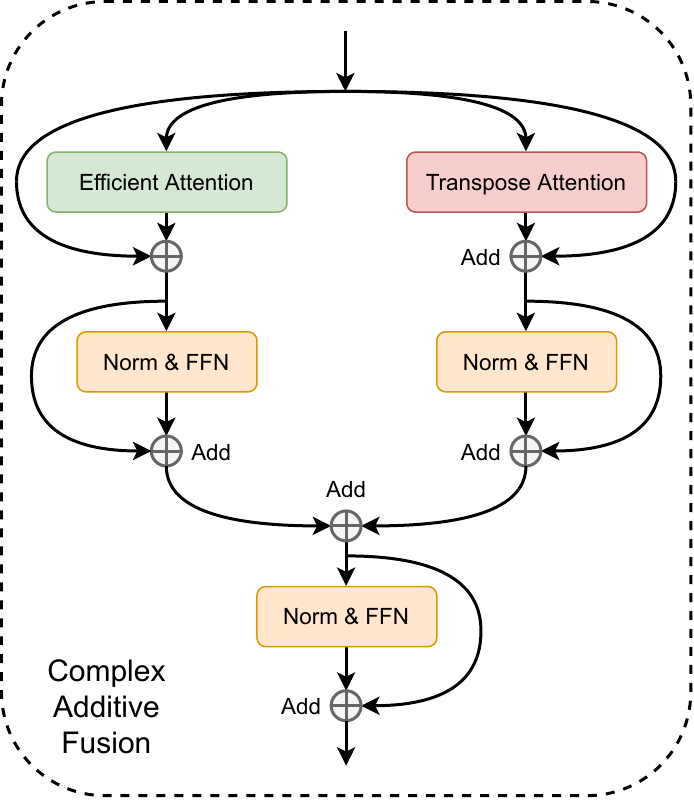}
  \end{subfigure}
  \begin{subfigure}
    \centering
    \includegraphics[width=0.45\linewidth]{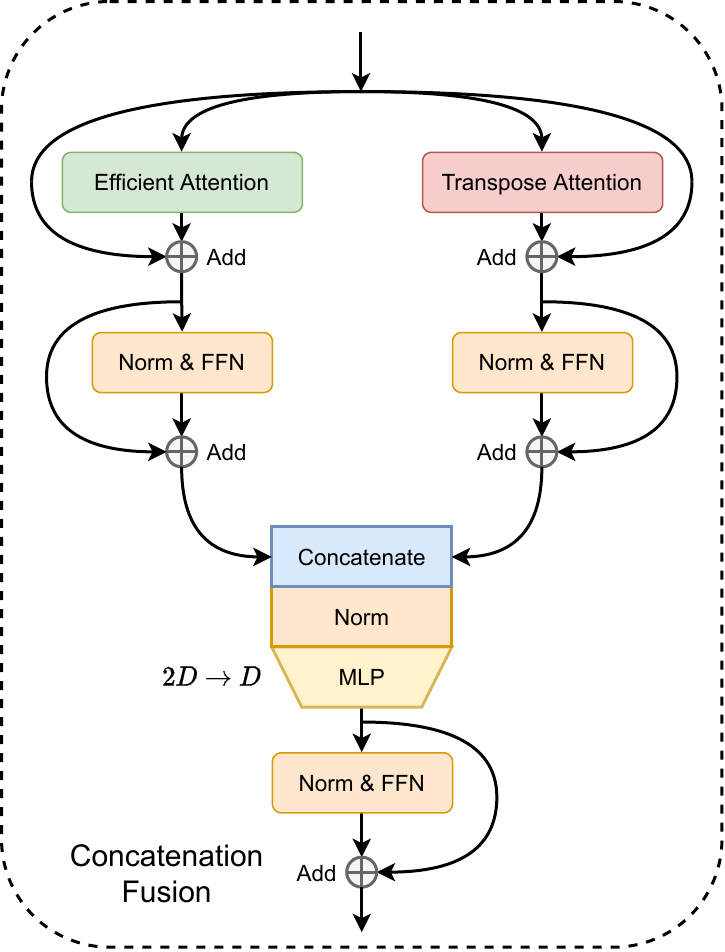}
  \end{subfigure}
  \caption{(a) Sequential dual attention, (b) simple additive dual attention, (c) complex additive dual attention, and (d) concatenation dual attention.}\label{fig-dual-attention-ablation}
\end{figure}

Four different possibilities are explored in this ablation study. These structures are shown in~\figureref{fig-dual-attention-ablation}. The first is sequential dual attention, as already explained in the methods section.
Next, two variants for additive dual attention were tested; namely, simple additive dual attention (\figureref{fig-dual-attention-ablation}(b)) and complex additive dual attention (\figureref{fig-dual-attention-ablation}(c)). The former was expected to have a more unstable backpropagation as the outputs of each attention block are not normalized. Although the latter has a larger number of parameters, the outputs of efficient and channel attention were normalized and fed to an FFN before the addition.
Lastly, the concatenation dual attention is shown in \figureref{fig-dual-attention-ablation}(d). The outputs of both blocks are normalized, then concatenated. An MLP reduces the dimension from twice the input dimension back to the input dimension, and the resulting tensor is again normalized.

The performance of the described dual attention modules was tested on the Synapse dataset. The network used is the same as shown in~\figureref{fig:proposed_method}, but the dual attention module is replaced by the modified versions.
The initial learning rate was set to 0.05 and the batch size to 24. A comparison is shown in~\tableref{tab-dual-attention-ablation}.
\begin{table}[h]
  \centering
  \begin{tabular}{ l  c  c  c }
    \toprule
    \textbf{Dual Attention Strategy}  & \textbf{\# Params [M]} & \textbf{DSC} & \textbf{HD}    \\
    \midrule
    Sequential       & 48.1  & 82.63            & 17.46 \\
    Simple Additive & 48.0  & 79.51            & 23.83 \\
    Complex Additive & 61.1  & 81.49            & 19.36 \\
    Concatenation    & 64.0  & 80.11            & 27.20 \\
    \bottomrule
  \end{tabular}
  \caption{Performance comparison of dual attention variations on the Synapse dataset.}\label{tab-dual-attention-ablation}
\end{table}
As can be seen, sequential fusion outperforms all the variants, with both the lowest required parameters and the best scores. Hence, we used it for all layers in the proposed network.

\section{Input Resolution and Skip Connection Effect}
We also explore the effect of skip connections in our suggested network. In this respect, we reconstructed our model with three settings: no skip connections, one skip connection in the highest layer, and two skip connections, i.e. the base setting.
The results are presented in~\tableref{tab-skip-conn}. We observe that the model behaves as expected: The more skip connections there are, the better the performance. In particular, fine-grained information from higher layers is critical to the fusion of information.

\begin{table}[!h]
    \centering
    \begin{tabular}{ c  c  c  c  c }
        \toprule
        \textbf{\# Skip Connections}      & \textbf{DSC} & \textbf{HD}    \\
        \midrule
        0                       & 79.71           & 22.46 \\
        1                       & 81.81           & 21.66 \\
        2                       & 82.63           & 17.46 \\
        \bottomrule
    \end{tabular}
    \caption{Performance comparison for different numbers of skip connections. Experiments are done on the Synapse dataset.}\label{tab-skip-conn}
\end{table}

To evaluate the effect of image resolution, we conducted experiments with two different image resolutions besides the original $224 \times 224$ resolution that we use in our experiments. In this respect, we reduced the input size to $128 \times 128$ pixels to observe the low-resolution impact. Furthermore, we slightly increased the image resolution to $288 \times 288$ pixels to analyze the effect of higher resolution. The results are shown in~\tableref{tab-image-size}. As expected, the performance increases for high resolution, as more fine-grained information is available.
\begin{table}[!h]
    \centering
    \begin{tabular}{ l  c  c  c }
        \toprule
        \textbf{Image Size} & \textbf{DSC} & \textbf{HD} \\
        \midrule
        $128\times 128$  & 79.20 & 15.90 \\
        $224\times 224 ^\dagger$  & 82.43  & 17.46 \\
        $288\times 288$ & 82.68  & 17.85 \\
        \bottomrule
    \end{tabular}
    \caption{Performance comparison for different image size settings. The symbol $\dagger$ indicates the original setting through this study.}\label{tab-image-size}
\end{table}

\section{Further Visualization Results}
\label{skin-results}
\begin{figure}[h]
\centering
\resizebox{\textwidth}{!}{
\begin{tabular}{@{} *{6}c @{}}
\includegraphics[width=0.16\textwidth]{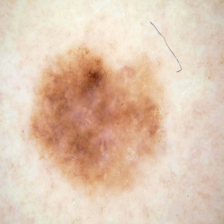} &
\includegraphics[width=0.16\textwidth]{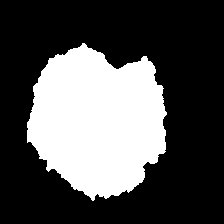} &
\includegraphics[width=0.16\textwidth]{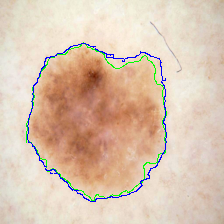} &
\includegraphics[width=0.16\textwidth]{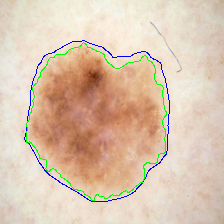} &
\includegraphics[width=0.16\textwidth]{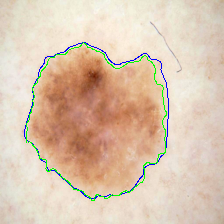} &
\includegraphics[width=0.16\textwidth]{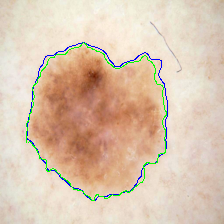} \\
\includegraphics[width=0.16\textwidth]{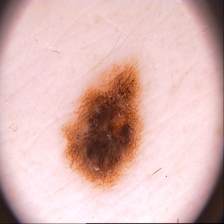} &
\includegraphics[width=0.16\textwidth]{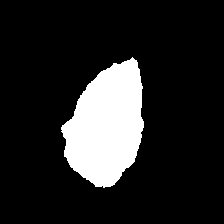} &
\includegraphics[width=0.16\textwidth]{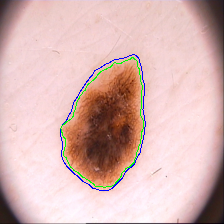} &
\includegraphics[width=0.16\textwidth]{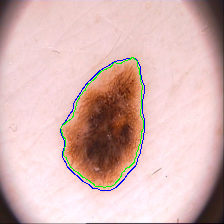} &
\includegraphics[width=0.16\textwidth]{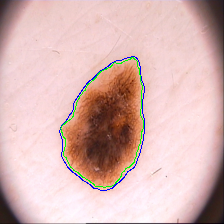} & 
\includegraphics[width=0.16\textwidth]{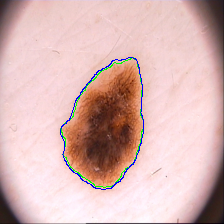} \\
{\small Input Image} & {\small Ground Truth} & {\small U-Net} & {\small Swin Unet} & {\small HiFormer} & {\small DAE-Former}
\end{tabular}
}
\caption{Visual comparisons of different methods on the \textit{ISIC2018} skin lesion dataset. Ground truth boundaries are shown
in \textcolor{green}{green}, and predicted boundaries are shown in \textcolor{blue}{blue}.} \label{fig:visualcomparison_isic}
\end{figure}

To further provide an insight into the segmentation capacity of the network, we provided comparative visualization results of the network on the skin lesion segmentation task in \figureref{fig:visualcomparison_isic}. It can be observed that our method estimates the skin lesion regions with varying shapes and patterns. Moreover, compared to the SOTA approaches, our method has a better prediction of the lesion boundary. 

\section{Computational Complexity}
The number of parameters used by the \transformershort is shown in \tableref{tab:parameters_synapse}.
As can be seen, although the \transformershort has a comparable number of parameters to the MISSFormer, it has a slightly better performance.

\begin{table*}[!ht]
    \centering
    \caption{Comparison of the number of parameters.}
    \begin{tabular}{l|c | c | c}
        \toprule
        \textbf{Methods}                        & \textbf{\# Params [M]} & \textbf{DSC~$\uparrow$} & \textbf{HD~$\downarrow$}
        \\
        \midrule
        DeepLapv3+ (CNN) \cite{chen2018encoder} & 59.50                  & 77.63                   & 39.95                    \\
        Swin-Unet \cite{cao2021swin}            & 27.17                  & 79.13                   & 21.55                    \\
        TransUNet \cite{chen2021transunet}      & 105.28                 & 77.48                   & 31.69                    \\
        LeVit-Unet \cite{xu2021levit}           & 52.17                  & 78.53                   & 16.84                    \\
        MISSFormer \cite{huang2021missformer}   & 42.5                   & 81.96                   & 18.20                    \\
        HiFormer \cite{heidari2022hiformer}   & 25.51                   & 80.39                   & 14.70                    \\
        \rowcolor{cyan!10}
        \transformershort                     & 48.1                   & 82.63                   & 17.46                    \\
        \bottomrule
    \end{tabular}\label{tab:parameters_synapse}
\end{table*}

\begin{figure}[h]
    \centering
    \includegraphics[width=0.78\textwidth]{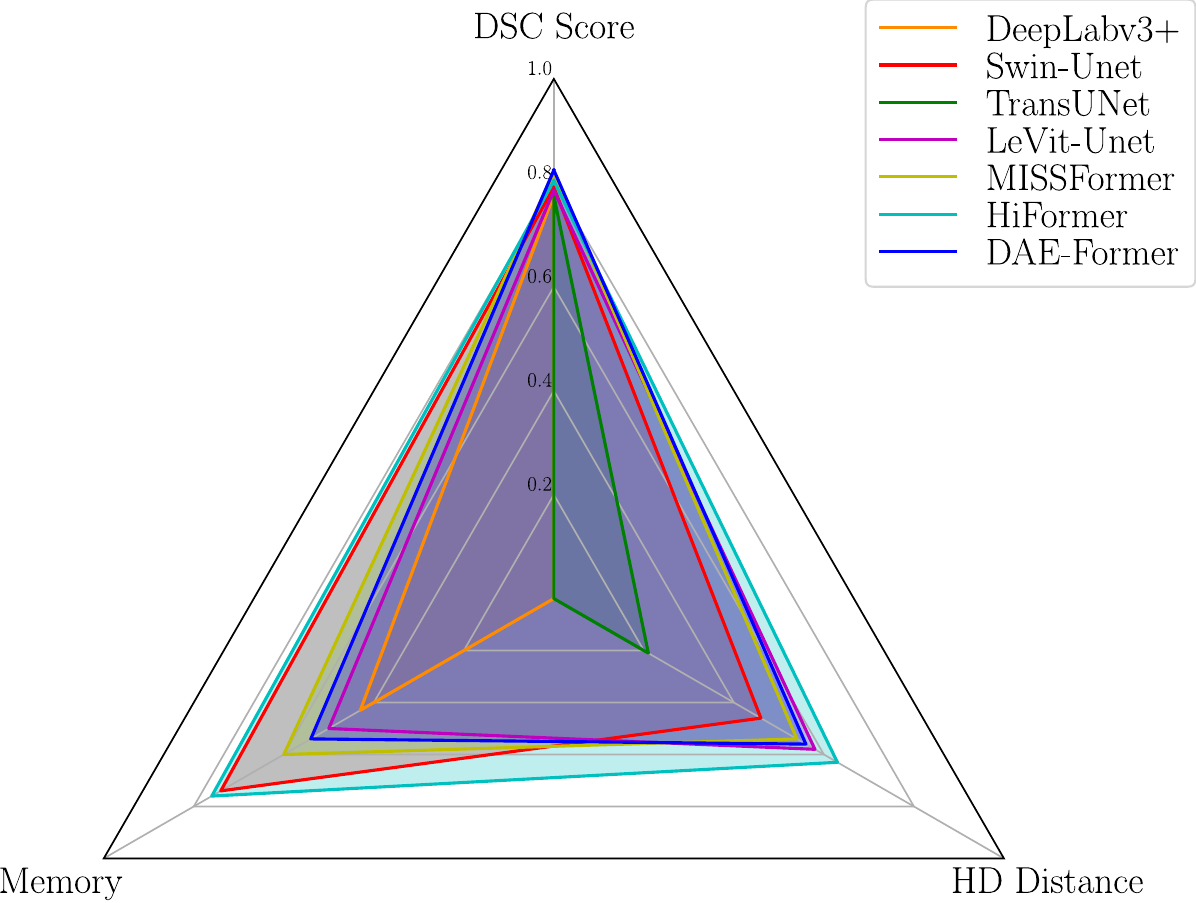}
    \caption{Efficiency vs performance chart of the SOTA approaches against our suggested network.}
    \label{fig:radar_chart}
\end{figure}

In order to compare the trade-off between efficiency and performance, we provided \figureref{fig:radar_chart}. In our visualization, we normalized each metric into the range 0-1, and then reversed the HD distance and the number of parameters to match the DSC score. It can be observed that our network performs better compared to other models with considerably lower parameters. TransUNet \cite{chen2021transunet}, for instance, requires 105.28M parameters, and its value is close to zero, whereas ours uses 48.1M parameters and achieves a significantly better Dice score and HD distance with 4.95 and 14.23 differences, respectively.

\let\cleardoublepage\clearpage
\end{document}